\documentclass[letterpaper, 10 pt, conference]{ieeeconf}

\IEEEoverridecommandlockouts                              

\usepackage{bm}
\usepackage{listings}
\usepackage{float}
\usepackage[dvipdfmx]{graphicx}
\usepackage{amsmath}
\usepackage{amssymb}
\usepackage{latexsym}
\usepackage{amsfonts}
\usepackage{url}
\usepackage{comment}
\usepackage{algorithm}
\usepackage{algpseudocode}

\newcommand{\FIG}[3]{
\begin{minipage}[b]{#1cm}
\begin{center}
\includegraphics[width=#1cm]{#2}\vspace*{-1mm}\\
{\scriptsize #3}\vspace*{3mm}~\\
\end{center}
\end{minipage}
}

\begin{document}


\newcommand{\CO}[1]{}

\newcommand{\editage}[2]{#1}

\newcommand{\figD}{
\begin{figure}[t]
  \centering
\FIG{8}{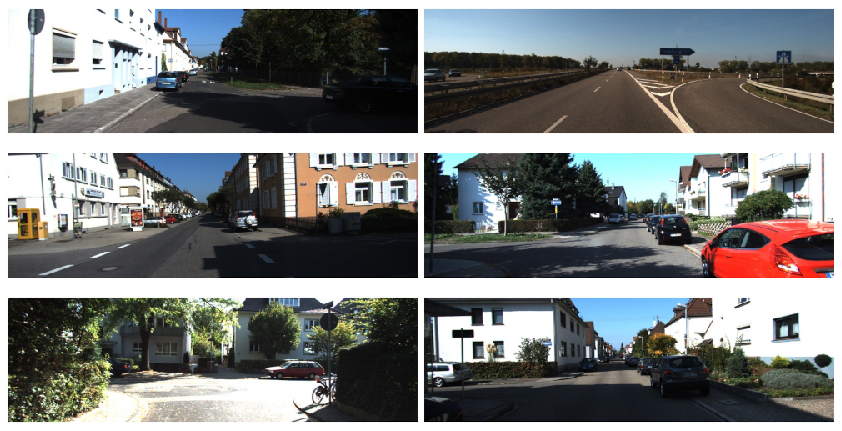}{}
  \caption{Examples of views corresponding to [-5, -10] m from the intersection location.}
\label{fig:pictures}
\end{figure}

}

\newcommand{\figF}{
\begin{figure}[t]
\begin{minipage}{8.5cm}
\FIG{4}{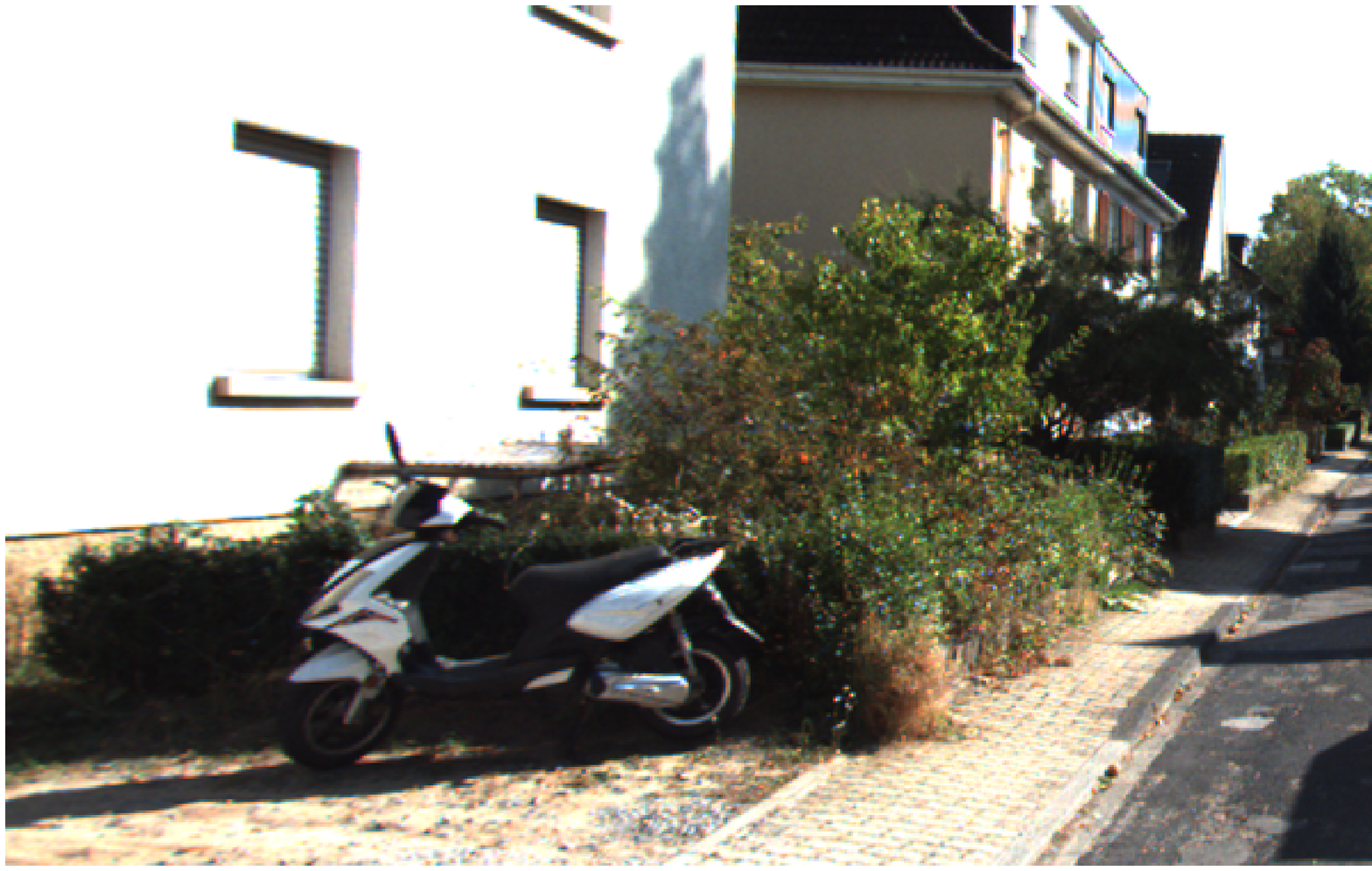}{go straight (class \#1)}
\FIG{4}{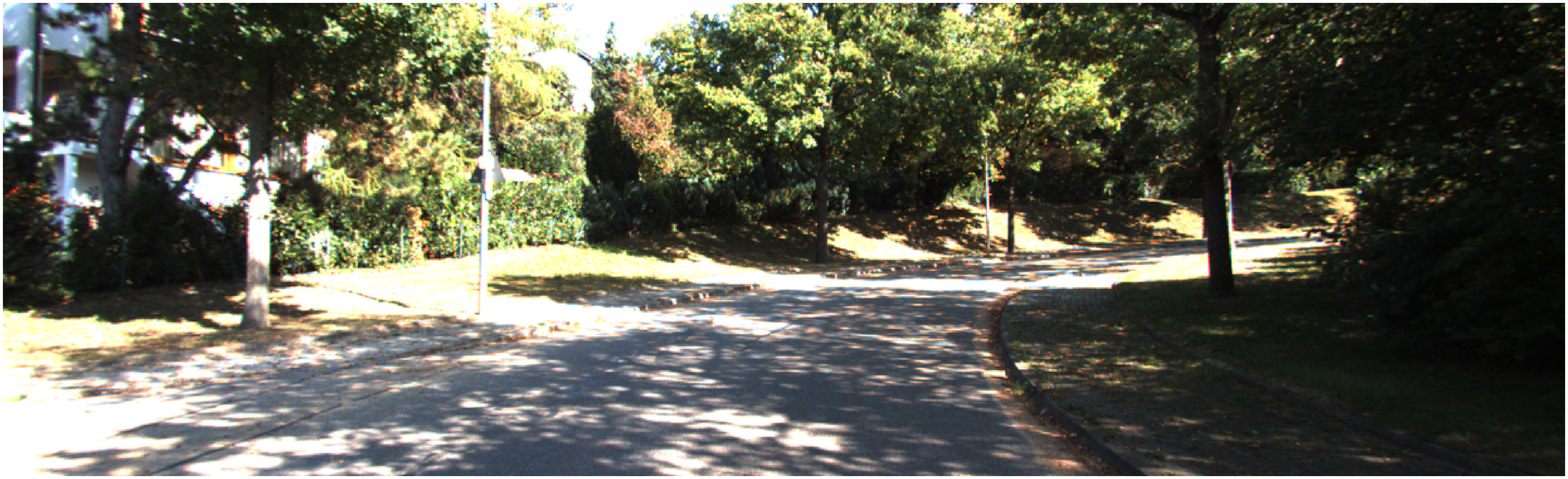}{turn right (class \#2)}
\FIG{4}{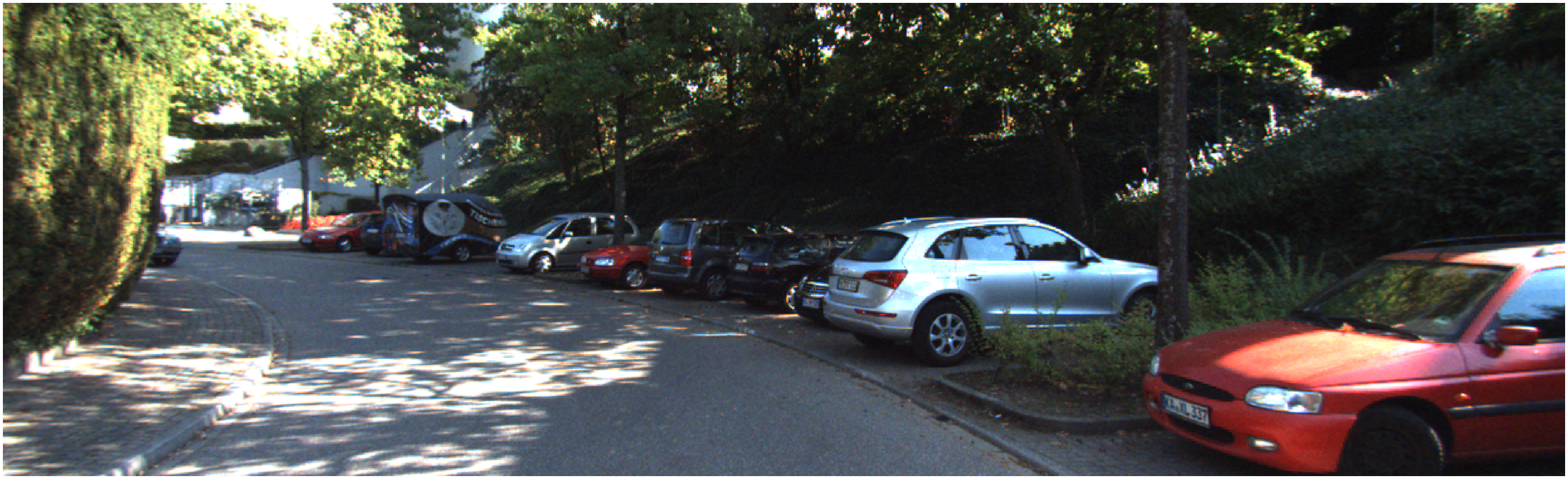}{turn left (class \#3)}
\FIG{4}{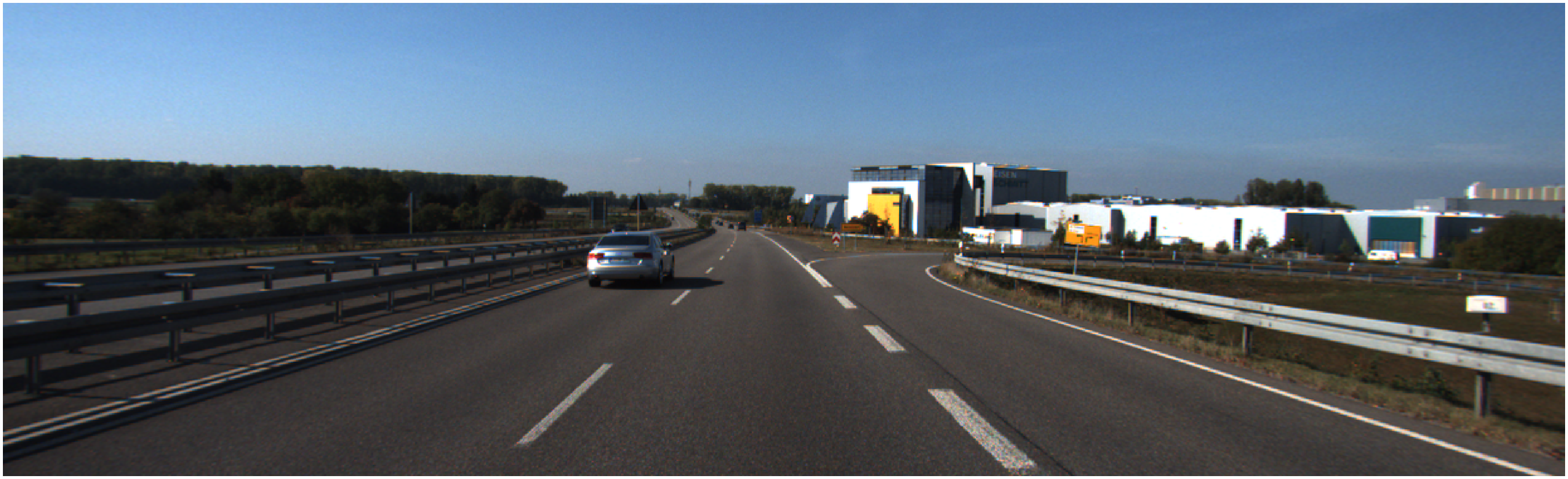}{right facing T-junction (class \#4)}
\FIG{4}{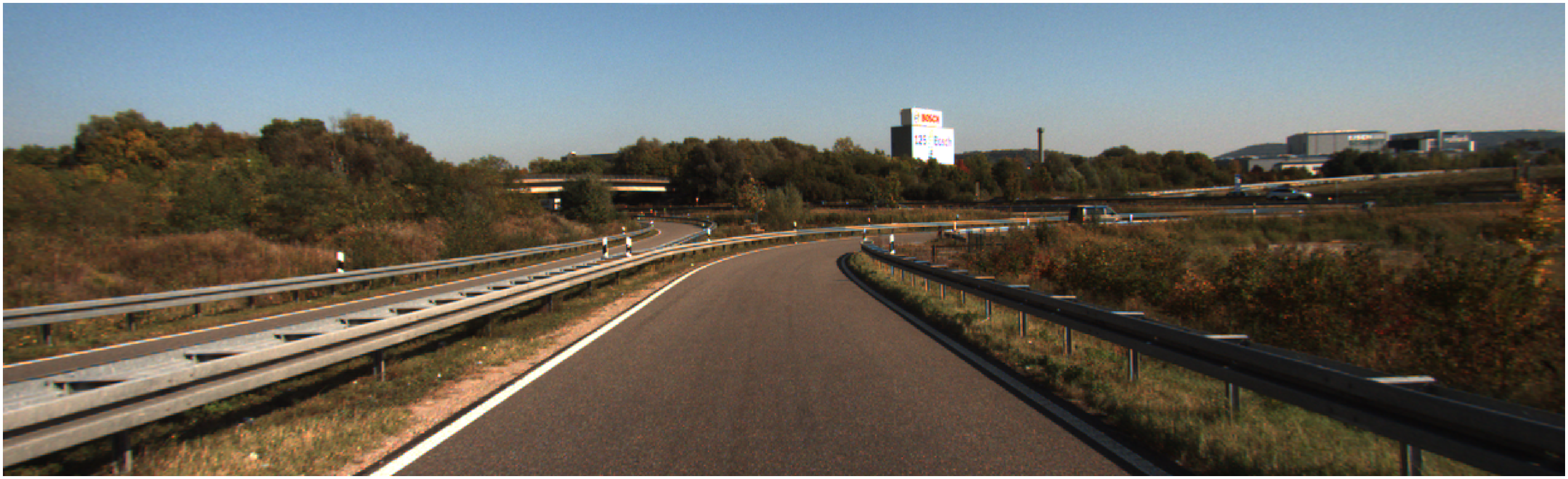}{turn right (class \#2)}
\FIG{4}{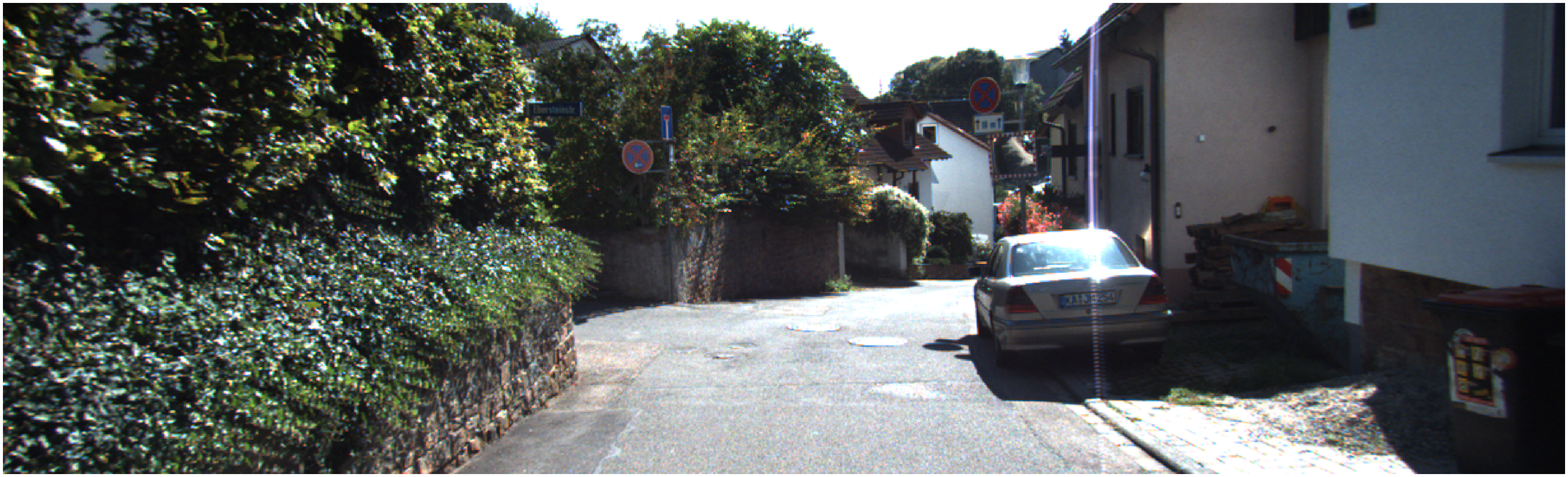}{bottom facing T-junction (class \#6)}
\FIG{4}{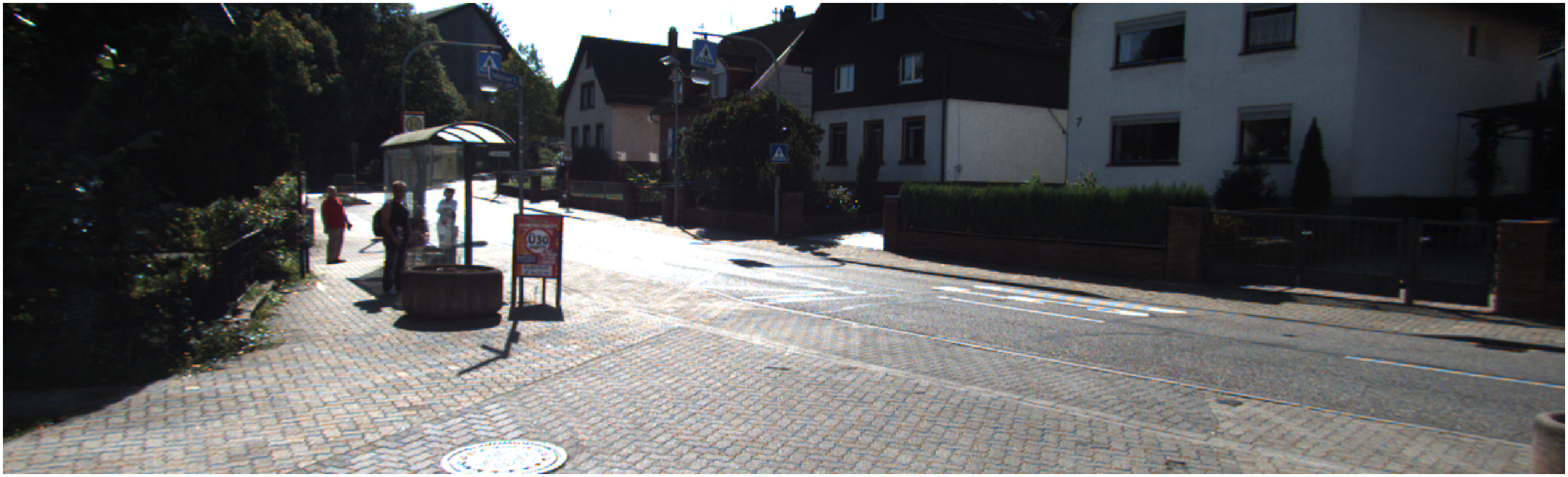}{bottom facing T-junction (class \#6)}
\end{minipage}
  \caption{Examples of input images}\label{fig:seven}
\end{figure}
}

\newcommand{\figK}{
\begin{table}[t]
\centering
\caption{Performance results}
\begin{tabular}{|r|r|}\hline
method&top-1 accuracy \\ \hline
TPV & 0.57 \\ \hline
FPV & 0.31 \\  \hline
VGG16 & 0.52 \\ \hline
SIFT+NBNN & 0.16 \\ \hline
LCF+NBNN & 0.23 \\ \hline
AE+L2 & 0.22 \\ \hline
Ours & {\bf 0.59} \\ \hline
\end{tabular}
\label{table:result}
\end{table}
}

\newcommand{\sss}{\hspace*{8mm}}

\newcommand{\figL}{
\begin{figure}[t]
\FIG{8}{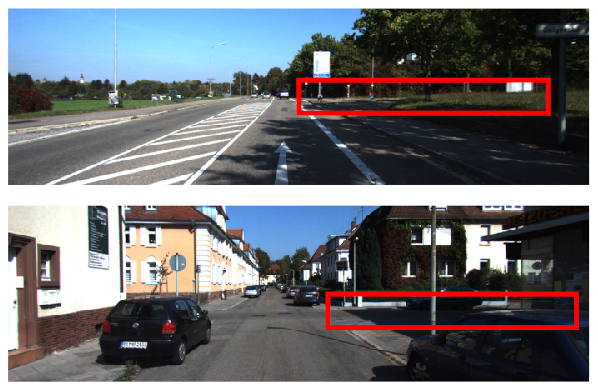}{(a)}\\
~\\
\FIG{8}{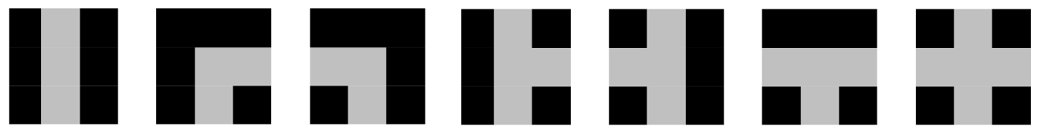}{}\\
{\scriptsize 
\hspace*{3mm}
\#1 \sss
\#2 \sss
\#3 \sss
\#4 \sss
\#5 \sss~
\#6 \sss
\#7 }\\
{\scriptsize \hspace*{3.5cm} (b)}\\
\caption{We aim to exploit the self-attention mechanism within the TPV module in the intersection classification system. (a) Motivative examples: Since most parts of the local pattern (e.g., road edges, buildings, and sky) are similar to each other, the ability of self-attention that can capture non-local context (e.g., the angle between two diagonal corners around an intersection would be effective). (b) The 7-class intersection classification problem, in which each class corresponds to 
\#1 go straight, 
\#2 turn right, 
\#3 turn left, 
\#4 right facing T-junction, 
\#5 left facing T-junction, 
\#6 bottom facing T-junction, and 
\#7 crossroad as shown in the figures.}\label{fig:7class}
\end{figure}
}

\newcommand{\figM}{
\begin{figure*}[t]
  \centering
\FIG{17}{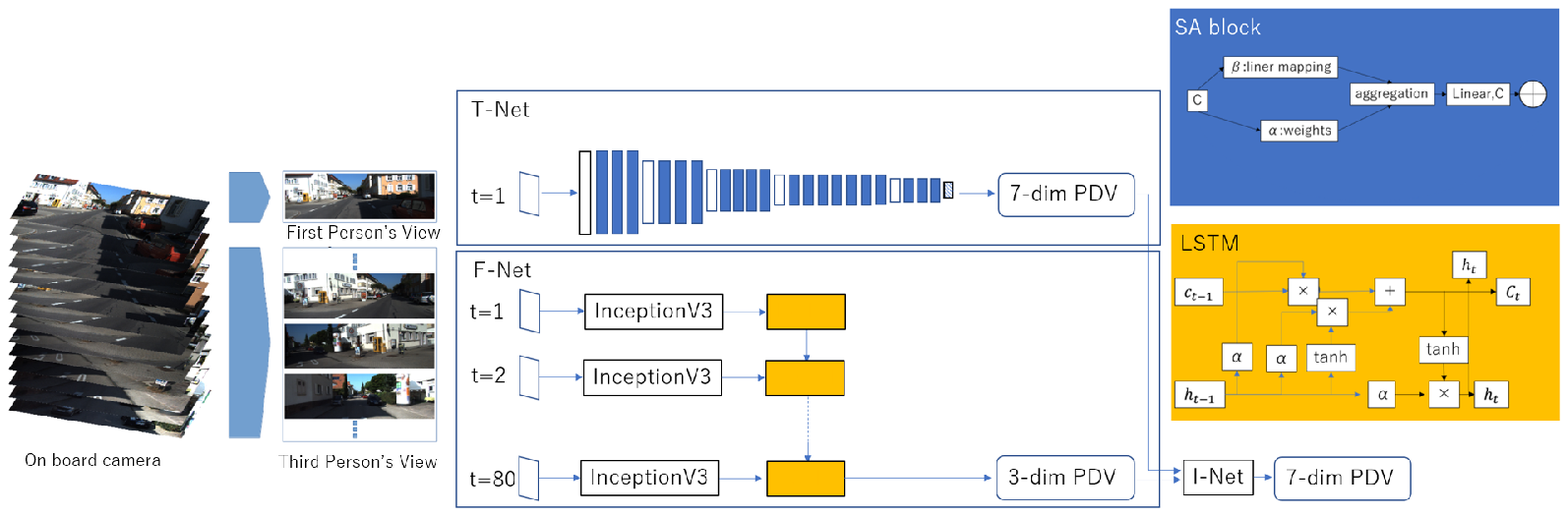}{}
  \caption{System overview.}\label{fig:overview}
\end{figure*}
}

\newcommand{\figP}{
\begin{figure*}[t]
\begin{center}
 \begin{minipage}{8cm}
  \begin{center}
\FIG{8}{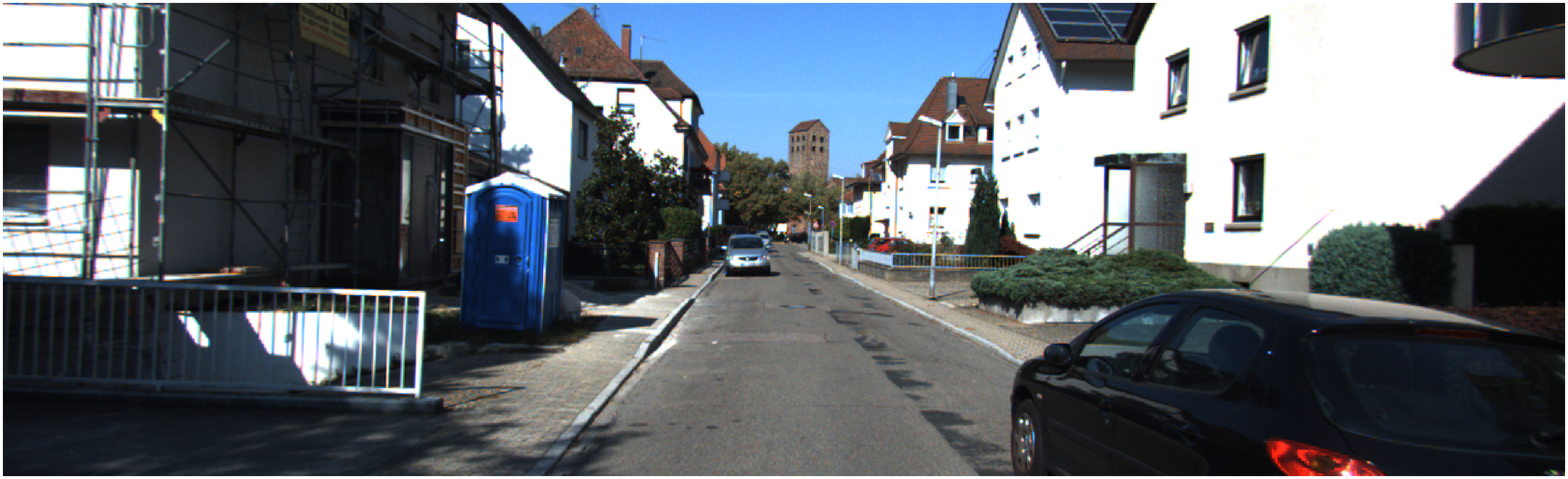}{\#1}
\FIG{8}{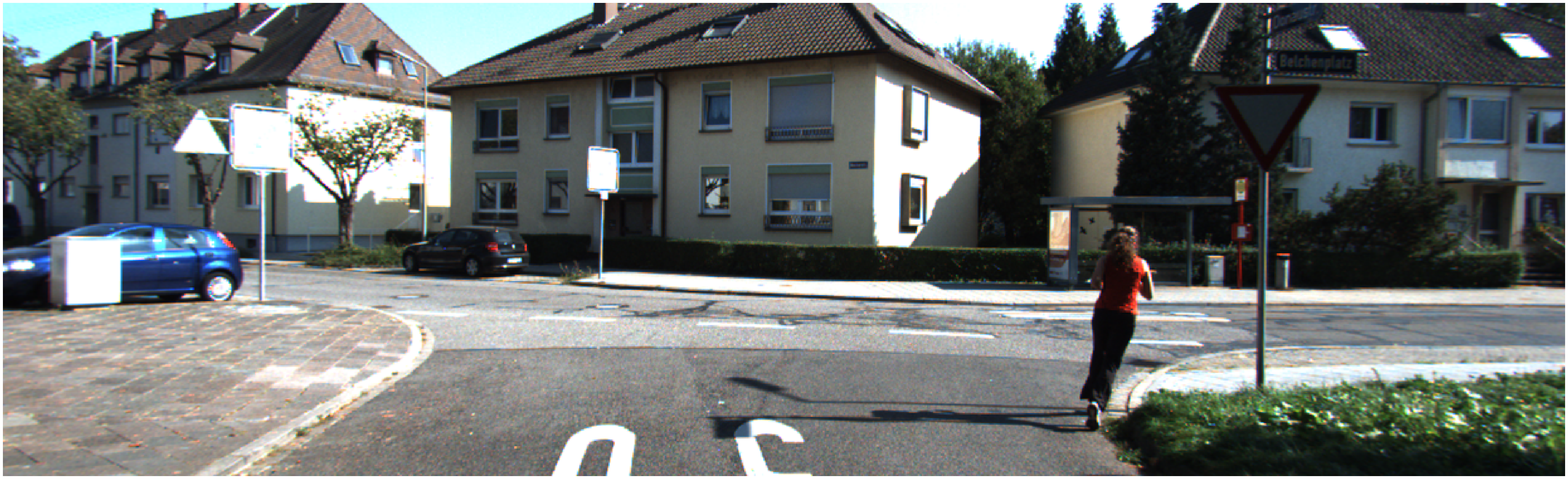}{\#2}
\FIG{8}{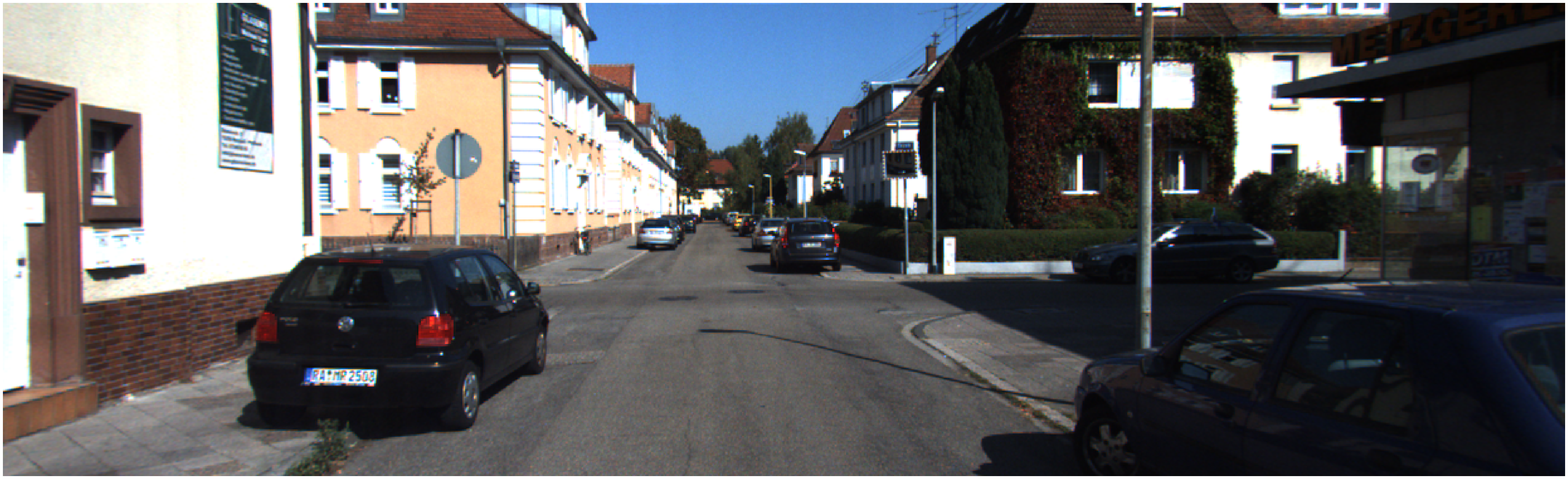}{\#3}
{\scriptsize (a) Success examples}
  \end{center}
 \end{minipage}
 \begin{minipage}{8cm}
  \begin{center}
\FIG{8}{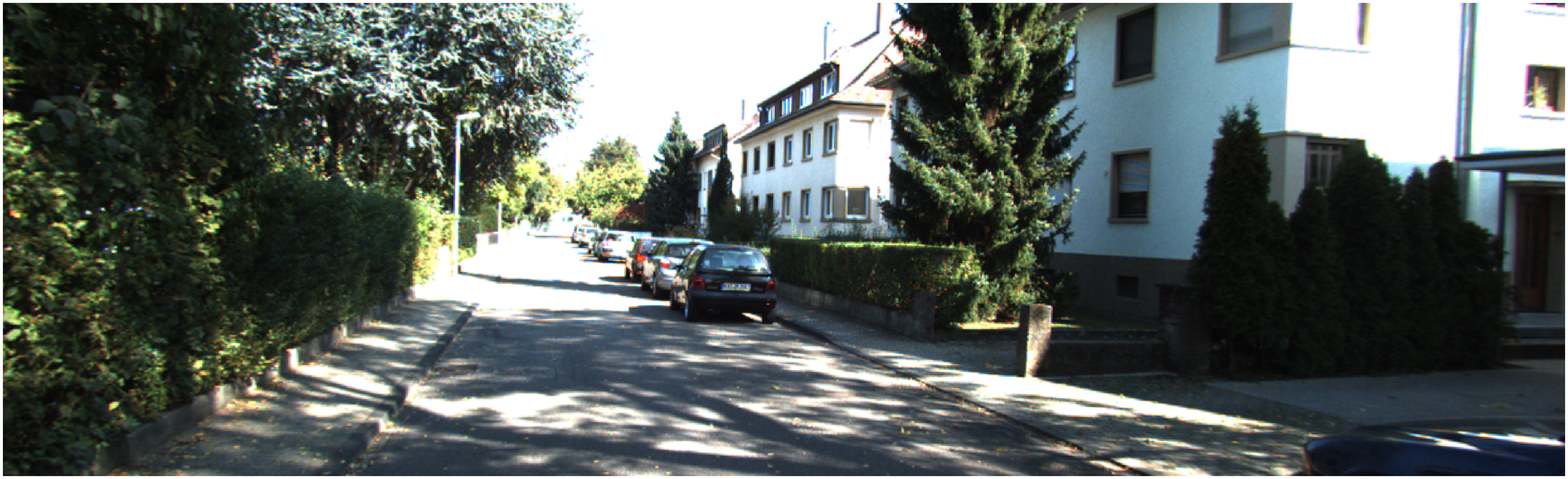}{\#1}
\FIG{8}{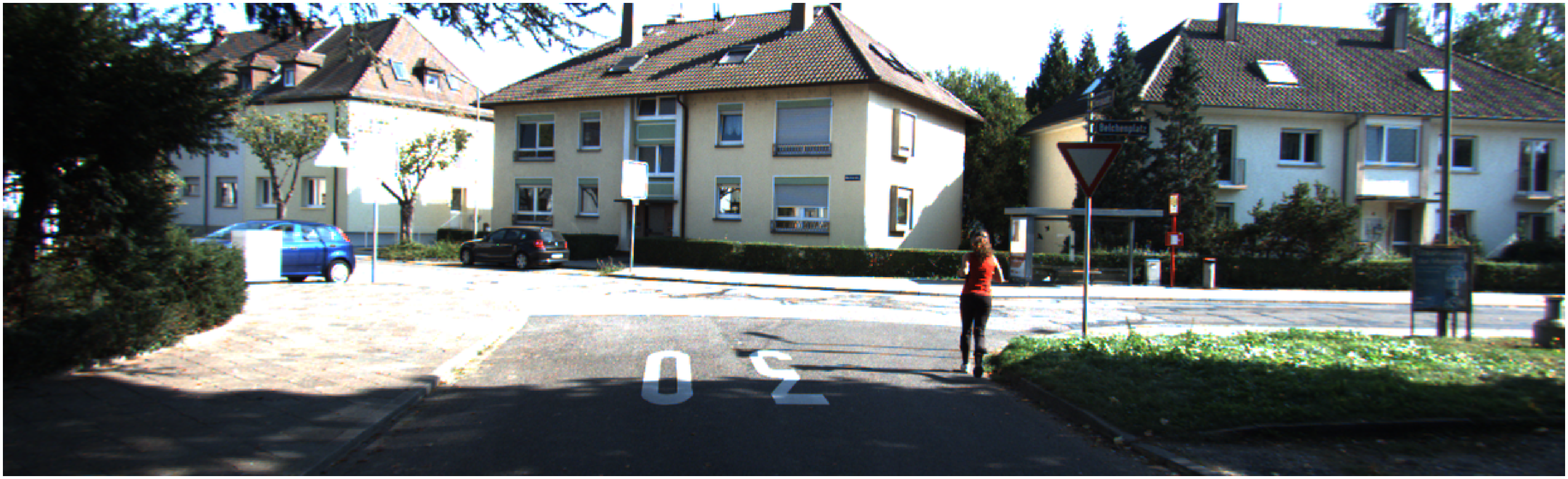}{\#2}
\FIG{8}{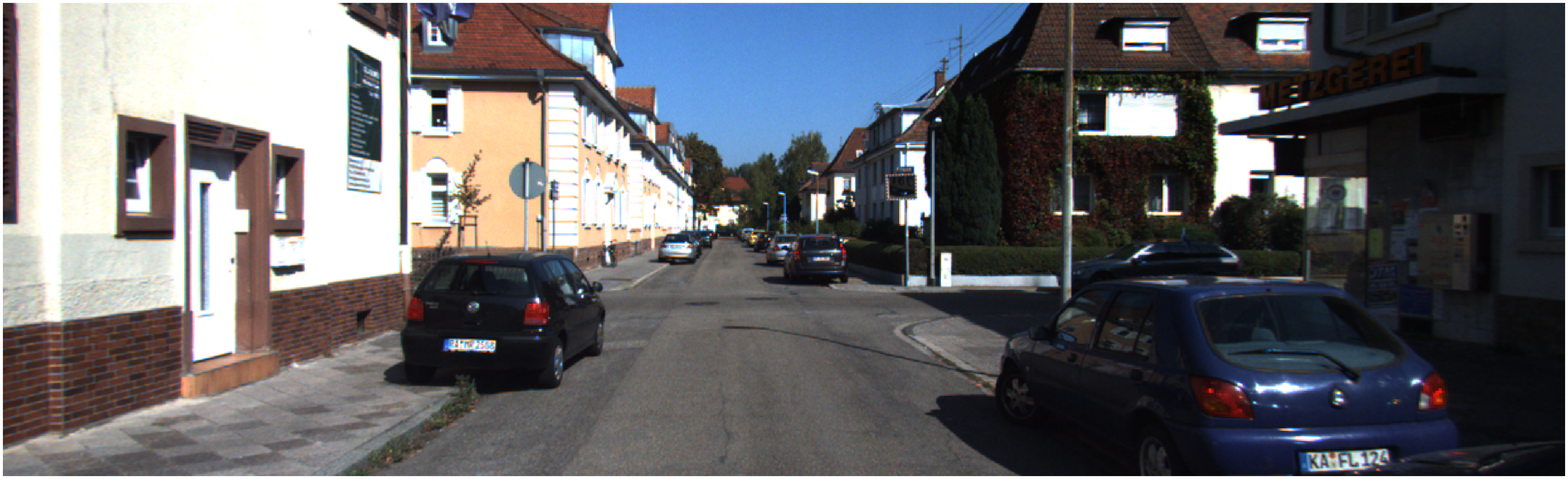}{\#3}
{\scriptsize (b) Failure examples}
  \end{center}
 \end{minipage}
\end{center}
\caption{Example results of TPV module.}\label{fig:successfailure}
\end{figure*}
}

\newcommand{\figS}{
\begin{figure}[t]
  \centering
\FIG{8}{imagenet_vgg16.eps}{}
  \caption{vgg16}
\label{fig:vgg16}
\end{figure}

}

\newcommand{\figT}{
\begin{figure}[t]
\FIG{8}{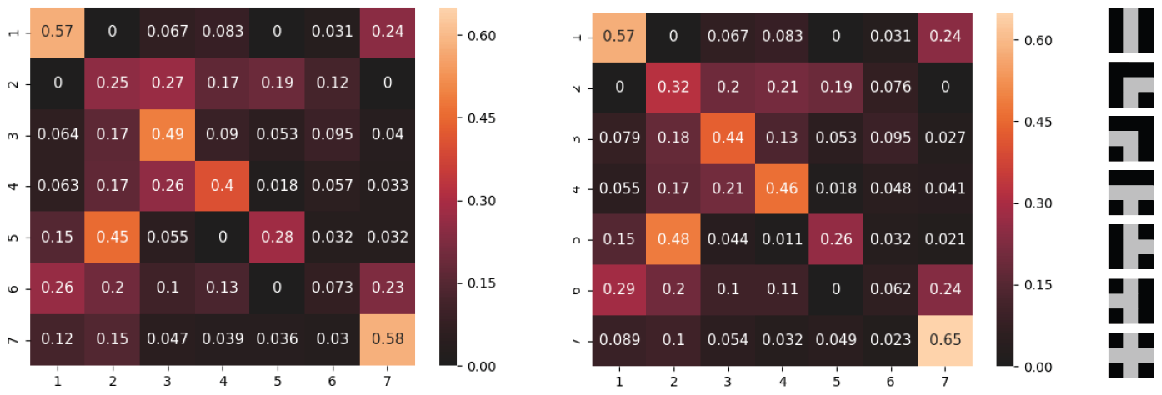}{}\\
\hspace*{1cm}{\scriptsize (a) TPV \hspace*{3cm} (b) FPV-TPV } \\
  \caption{Confusion matrix.}\label{fig:confusion}
\end{figure}

}

\title{\LARGE \bf
Exploring Self-Attention for Visual Intersection Classification
}

\author{%
Haruki Nakata, Kanji Tanaka, and Koji Takeda
\thanks{Our work has been supported in part by 
JSPS KAKENHI 
Grant-in-Aid 
for Scientific Research (C) 17K00361 
and 20K12008.}
\thanks{$*$H. Nakata, K. Tanaka, and K. Takeda are with Department of Engineering, University of Fukui, Japan. {\tt\small tnkknj@u-fukui.ac.jp}}
}

\maketitle

\begin{abstract}
In robot vision, self-attention has recently emerged as a technique for capturing non-local contexts. In this study, we introduced a self-attention mechanism into the intersection recognition system as a method to capture the non-local contexts behind the scenes. An intersection classification system comprises two distinctive modules: (a) a first-person vision (FPV) module, which uses a short egocentric view sequence as the intersection is passed, and (b) a third-person vision (TPV) module, which uses a single view immediately before entering the intersection. The self-attention mechanism is effective in the TPV module because most parts of the local pattern (e.g., road edges, buildings, and sky) are similar to each other, and thus the use of a non-local context (e.g., the angle between two diagonal corners around an intersection) would be effective. This study makes three major contributions. First, we proposed a self-attention-based approach for intersection classification using TPVs. Second, we presented a practical system in which a self-attention-based TPV module is combined with an FPV module to improve the overall recognition performance. Finally, experiments using the public KITTI dataset show that the above self-attention-based system outperforms conventional recognition based on local patterns and recognition based on convolution operations.
\end{abstract}

\section{Introduction}

In the existing scene recognition technology for autonomous driving applications, the convolution operator \cite{x} is often used to process local neighbor information using a fixed weight kernel to hierarchically aggregate the global context. However, this method makes it difficult to capture non-local contexts at a high spatial resolution, which is not sufficient for typical applications where a high inter-class similarity of local feature distributions is expected.

\editage{
\figL
}{}

Self-attention \cite{Paper:SAN} has recently emerged as a technique for capturing non-local contexts. An important concept of self-attention is to calculate the global context for a target region as a weighted sum of features over all image regions. The corresponding weights were dynamically calculated using the similarity function between the features of the embedded space at these locations. The number of parameters was independent of the scale at which self-attention handles long-distance interactions. This allowed us to capture non-local contexts with high spatial resolution.

In this study, we introduced a self-attention mechanism into the intersection recognition system as a method to capture the non-local contexts behind the scenes (Fig. \ref{fig:7class}a). The goal of intersection classification is to classify traffic scenes with respect to the road topology (e.g., a seven-class problem, as illustrated in Fig. \ref{fig:7class}b). An intersection classification system consists of two distinctive modules: (a) first-person vision (FPV), which uses a short egocentric view sequence as the intersection is passed; and (b) third-person vision (TPV), which uses a single view immediately before entering the intersection. The self-attention mechanism is effective in the TPV module because most parts of the local pattern (e.g., road edges, buildings, sky) are similar to each other, as shown in Fig. \ref{fig:successfailure}; thus, the use of a non-local context (e.g., the angle between two diagonal corners around an intersection) would be effective. Our idea was to exploit self-attention as an end-to-end learning method to capture long-distance contextual information.

This study makes three major contributions. First, we propose a self-attention-based approach for intersection classification using TPVs. Second, we present a practical system in which a self-attention-based TPV module is combined with an FPV module to improve the overall recognition performance. Finally, experiments using the public KITTI dataset show that the above self-attention-based system outperforms conventional recognition based on local patterns and recognition based on convolution operations.

\section{Related Work}

Intersection recognition is formulated as an image-classification problem. For example, \cite{Paper:segmentationbased} presented an approach based on a cascade of feature extractors and classification modules. More recently, deep learning approaches, such as the application of a deep convolutional neural network to this problem, have proven to be effective by several researchers. For example, \cite{Paper:Jurai} presented a multitask learning method that simultaneously performs intersection recognition and segmentation tasks. Other traffic scene recognition problems have also been studied extensively. For example, crossability prediction \cite{Paper:18} and the division/reconstruction of crossable road areas \cite{Paper:19}. In \cite{Paper:20}, a deep convolutional neural network-based road surface classifier was employed to distinguish between six road surface types. In \cite{Paper:22}, a data-adaptive similarity scale was used to cluster the traffic conditions. In \cite{Paper:23}, a large-scale input image was used for crack detection. In \cite{Paper:19}, a monocular detection method was proposed for robust road detection under severe occlusions. In \cite{Paper:25}, contextual information and traffic rules were considered to predict pedestrian movement. In \cite{Paper:9}, a stereo vision algorithm was used to estimate the scene layout by describing static/dynamic objects. In \cite{Paper:11}, this study was extended to detect intersections in a view-invariant manner.

\editage{
\figM
}{}

Intersection recognition is closely related to the problem of road topology recognition based on ego-motion predictions. In \cite{brubaker2015map}, map-based self-localization was achieved using the motion cues of a vehicle measured by visual odometry. In \cite{Paper:12}, a visual odometry method was developed to calculate the movement of a monocular camera with six degrees of freedom, assuming that the ground plane is known. In \cite{Paper:13}, a robust new visual odometry framework was presented by considering the motion model of the vehicle. In \cite{Paper:DeepVO}, an end-to-end deep learning framework was used to train a regressor for visual odometry. In \cite{Paper:14}, a hierarchical graph-based scene representation was used to understand complex urban environments. Other video-analysis problems have also been studied extensively. For example, in \cite{Paper:15}, lane change prediction was studied using LSTM. In \cite{Paper:16}, the performance of a lateral-position prediction algorithm was evaluated. In \cite{Paper:17}, a situation resulting from future lane changes was predicted. In \cite{Paper:24}, a probabilistic multisession framework was developed using high-precision road maps for urban areas using low-cost sensors. In \cite{Paper:10}, intersection detection was studied as a binary classification problem using LSTM based on a monocular sequence. In \cite{Paper:21}, a framework was developed for the simultaneous estimation of highway traffic conditions and traffic flow parameters.

Studies have also been conducted with additional cues from other modalities, such as global positioning systems (GPS) \cite{Paper:26}, publicly available map data \cite{Paper:27}, road signs \cite{Paper:28}, and domain-specific prior knowledge, such as traffic lights \cite{Paper:29}. Recent research examples include the classification and interpretation of signs that are not affected by the weather \cite{Paper:30}, detection of traffic lights and small objects \cite{Paper:31}, detection and recognition of signs \cite{Paper:28}, hierarchical detection of traffic lights \cite{Paper:32}, and cross-dataset classification of traffic lights \cite{Paper:33}. In addition, detecting and tracking moving objects and pedestrians to predict the road area where the vehicle can travel contributes to intersection recognition from another direction \cite{Paper:34}.

The proposed approach has three main advantages. First, to the best of our knowledge, we are the first to explore the self-attention approach in the context of visual intersection classification. Second, our framework achieves improved performance by combining the above self-attention-based TPV with the FPV approach for ego-motion prediction. Third, this framework consists of a minimal set of modules, FPV and TPV, which are complementary and orthogonal to the aforementioned methods based on other modalities.

\section{Proposed Approach}

We aim to boost the intersection classification system, which consists of FPV and TPV modules, by introducing a self-attention mechanism to capture the global context. Figure \ref{fig:overview}  presents an overview of the proposed framework. Specifically, the self-attention mechanism was applied to the TPV module (``T-Net" in Fig. \ref{fig:overview}), in which the 7-class intersection classification problem was directly addressed (Section \ref{sec:sa}). Subsequently, the FPV module (``F-Net" in Fig. \ref{fig:overview}) is introduced as a motion classifier. It classifies vehicle ego-motions into three major classes: ``go straight," ``turn right," and ``turn left". Thereafter, the FPV module is combined with the TPV module to improve the overall performance of intersection classification (Section \ref{sec:fpvtpv}). The output of either the FPV or TPV module is in the form of a class-specific probability density vector (PDV). Although the FPV module is implemented in the proposed method as a 3-class motion classifier, it can be implemented to address the 7-class intersection classification directly, which will be discussed in an ablation study in the experimental section (Section \ref{sec:exp}).

\subsection{Self-attention Mechanism}\label{sec:sa}

The self-attention mechanism was originally revolutionized extensively in areas, such as natural-language processing and machine translation. Since then, it has been applied in fields, such as image recognition, question answering, and image generation \cite{Paper:la}. However, until recently, it has been used to complement the convolution operations, adjust the output of the convolution, or create a layer that is used by combining the convolution with other methods. However, recent studies have shown that limiting self-attention to local patches can adapt it to the network and create better models in terms of robustness and generalization \cite{Paper:SAN}.

In \cite{Paper:SAN}, a method of using an image recognition model with a self-attention mechanism was proposed as a study of scene classification. It addresses two variants: patch-wise and pairwise self-attention. The patch-wise method can uniquely identify a particular location in a footprint. Pairwise generalizes the self-attention mechanism to standard dot product attention. In our implementation, patch-wise self-attention showed a higher performance empirically, and we decided to use it as the default method. In this case, self-attention takes the form
\begin{eqnarray}
y_{i}=\displaystyle\sum_{j\in R(i)} \alpha(x_{R_{(i)}})_j \odot \beta(x_j),
\end{eqnarray}
where $x_{R(i)}$ is a patch of the feature vector of footprint $R(i)$, and $\alpha(x_{R(i)})$ is a tensor with the same spatial dimension as the patch $x_{R(i)}$. $\alpha(x_{R(i)})_j$ is the vector at location $j$ in this tensor, and it corresponds to the vector $x_j$ in $x_{R(i)}$. Contrary	to a CNN, which can handle only nearby information, in self-attention, the weight calculation $\alpha(x_{R(i)})$ can be indexed individually and the feature vector $x_j$ can be indexed by location. This makes it possible to fuse the feature vector information from different locations in the footprint. $\alpha(x_{R(i)})$ is assumed to be decomposed by the equation:
\begin{equation}
\alpha(x_{R(i)})=y(\delta(x_{R(i)})).
\end{equation}
For
$\delta(x_{R(i)})$,
three different combinations 
as followings are explored:
\begin{itemize}
\item
Star-product:
\begin{equation}
\delta(x_{R(i)}) =\displaystyle [ \varphi(x_i)^T \psi(x_j) ]_{\forall j \in R(i)}
\end{equation}
\item
Clique-product:
\begin{equation}
\delta(x_{R(i)}) =\displaystyle [ \varphi(x_j)^T \psi(x_k) ]_{\forall j, k \in R(i)}
\end{equation}
\item
Concatenation:
\begin{equation}
\delta(x_{R(i)}) =\displaystyle [ \varphi(x_i), [ \psi(x_j) ]_{\forall j \in R(i)} ]
\end{equation}
\end{itemize}
The star product has small flops, but its accuracy is not satisfactory. Concatenation has the highest accuracy among the three types; however, it has large flops. The liquid product had a good balance between the two methods. 

In this study, we prioritized accuracy and used concatenation.

We pretrained the self-attention mechanism on ImageNet, a large dataset of 1,000 object classes, and then adapted the model to the target task via transfer learning. The fully connected layer of the final layer is modified into seven classes. Fine-tuning was then performed to adapt to a new task. For fine-tuning, the weights of the six layers, including the fully connected layer and transition layer in the architecture, were set learnable, and those of the other layers were prefixed.

Figure \ref{fig:overview} presents an overview of the TPV module (``T-Net"). As shown in the figure, the TPV module comprised five transition layers, 19 self-attention blocks, and one output layer. A CNN was used for feature extraction and conversion.

There are five transition layers, each output is 112$\times$112$\times$64, 56$\times$56$\times$256, 28$\times$28$\times$512, 14$\times$14$\times$1,024, and 7$\times$7$\times$2,048. By lowering the spatial resolution, the computational load is reduced and the receptive field is expanded. The transition consists of a batch normalization layer, ReLU \cite{Paper:42}, 2$\times$2 maximum pooling with stride 2, and a linear mapping that extends the channel dimension.

In the self-attention block (``SA block" in Fig. \ref{fig:overview}), the input feature tensor (channel dimension $C$) passes through two processing streams. The stream at the bottom evaluates the attention weight $\alpha$ by computing the function $\delta$ (via the mappings $\varphi$ and $\psi$) followed by the mapping $\gamma$. The stream at the top applies a linear transformation $\beta$ that transforms the input features and reduces their dimensions for efficient processing. The outputs of the two streams were aggregated using the Hadamard product. The combined features undergo normalization and elementwise nonlinearization and are then processed in the final linear layer, to expand the dimension to $C$.

\subsection{FPV-TPV Framework}\label{sec:fpvtpv}

Deep visual odometry in \cite{Paper:DeepVO} was employed as the basis for our F-Net (``F-Net" in Fig. \ref{fig:overview}). It combines the advantages of convolutional neural network and recursive LSTM neural network, and is one of the best-known methods for ego-motion estimation. 

Using raw sequence images as the input of the LSTM did not yield a good performance experimentally. We found that using an optical flow sequence instead of a raw sequence often significantly improves performance. Therefore, this optical flow-based variant was used in our study. Each pixel of an optical flow image is the intensity and direction calculated from the 2-channel array containing the optical flow vector $(u, v)$ \cite{Paper:Farneback}, and is then color-coded.

The FPV module processes the features of each frame using LSTM (``LSTM" in Fig. \ref{fig:overview}). It extracts 2,048-dimensional features from the feature extraction part of the convolutional neural network of Inception V3 for each frame. This feature vector is used as the input for the LSTM for each time step. Thereafter, for the feature extraction part using Inception V3, feature extraction was performed using the weights learned by ImageNet as they were. Subsequently, the parameters of the FPV module were updated by setting only the LSTM and fully connected layer to be learnable.

In this study, the posterior distribution of each class was obtained using the output of the FPV module as a prior distribution and expanding the output of the TPV module (``I-Net" in Fig. \ref{fig:overview}). The output of the I-Net is calculated as $I[c] = W_S[c] P_M[c]$ for each class $c$. Furthermore, when the confidence of top-1 of the FPV module is 0.9999 or higher, the 7-dimensional mask vector $T$ is introduced, and $I[c] = W_S[c] P_M[c] T[c] $. $W_S$ assumes one of the three values depending on the results of the FPV module. This is controlled by the ego-motion class $c^{-}$ ($\in$ \{ ``go straight," ``turn right," ``turn left" \}) that received the lowest likelihood value:
If $c^{-}$ is ``go straight":
\[
\bm{W_S} = (0, 1, 1, 1, 1, 1, 1),
\]
or if $c^{-}$ is ``turn right":
\[
\bm{W_S} = (1, 0, 1, 1, 1, 1, 1)
\]
or if $c^{-}$ is ``turn left":
\[
\bm{W_S} = (1, 1, 0, 1, 1, 1, 1)
\]

In addition, the mask vector $T$ is controlled by the ego-motion class $c^{+}$ ($\in$ \{ ``go straight," ``turn right," ``turn left" \}) that received the highest likelihood value: If $c^{+}$ is ``go straight":
\[
\bm{T} = (1, 0, 0, 0, 1, 1, 1)
\]
or if $c^{+}$ is ``turn right":
\[
\bm{T} = (0, 1, 0, 1, 1, 0, 1)
\]
or if $c^{+}$ is ``turn left":
\[
\bm{T} = (0, 0, 1, 1, 0, 1, 1)
\]

\section{Evaluation Experiments}\label{sec:exp}

The KITTI dataset was used to experimentally evaluate the proposed method and other comparative methods.

\subsection{Comparing Methods}

Six comparing methods,~TPV,~FPV,~VGG16,~SIFT+NBNN,
LCF+NBNN,
and
AE+L2
were employed for performance comparison.
(1)
The TPV method uses a T-Net based on the self-attention mechanism, as described in Section \ref{sec:sa}.
(2)
The FPV method is based on almost the same architecture as F-Net, but the output layer is modified to address the 7-class intersection classification directly, not the 3-class motion classification.
(3)
The VGG16 method is a convolutional neural network consisting of 13 convolutional layers and three fully connected layers, for a total of 16 layers \cite{Paper:VGG16}. In its training stage, publicly available weights pretrained on ImageNet were further finetuned to address the 7-class problem. 
(4)
SIFT+NBNN
is based on the scene representation of 
a bag of 128-dim SIFT features \cite{takeda44} with Harris-Laplace keypoints (1,500-2,000 per image) using the naive Bayes nearest neighbor (NBNN) distance metric \cite{takeda45}.
(5)
LCF+NBNN
is 
different from SIFT+NBNN
only in that the
512-dim LCF feature \cite{takeda46} (768 per image) 
is 
used instead of the 
128-dim SIFT. 
(6) 
AE+L2
is based on 
a global 3,136-dim autoencoder (AE) feature with the nearest neighbor-based distance metric \cite{takeda47}. 
Our decision to use the NBNN distance metric was motivated by its success in previous studies on visual place recognition \cite{takeda48}. We used L2-norm to measure the distance between a feature pair for all the methods.

\editage{
\figD

\figF

\figK

\figP

\figT
}{}

\subsection{Settings}

In this study, a new dataset was created by extracting the entire intersection image and the intersection passage sequence from the KITTI data set \cite{KITTI}. The image from the left eye of the stereo camera of the KITTI dataset was used as the input for our system. The image size is 241$\times$376 pixels. The input layer of the T-Net was slightly modified to address our image size of 224$\times$224 pixels. The input images of the FPV module derived from Inception V3 were resized to 299$\times $299 pixels to be fitted to the network.
To stabilize the training process, the training sequences for F-Net were clustered in terms of the time interval and clusters whose size is smaller than five were not used for training.

The details of the training are as follows. A GeForce RTX 3090 GPU is used to train the TPV and FPV modules. A PyTorch virtual environment and Keras with a TensorFlow backend were used for the F-Net and I-Net. The number of epochs was set as 20.

The details of dataset creation are as follows. We manually analyzed the GPS information and obtained an image corresponding to the range $[-L_2, -L_1]$ from the center of the intersection. The hyper-parameters are set as $L_1$=0 [m], and $L_2$=5 [m] following \cite{Paper:laaa}. From these images, 152 images were extracted for each of the seven classes. An example of an image of the KITTI dataset corresponding to this interval is shown in Fig. \ref{fig:pictures}. The image set was divided into 126 training datasets and 36 test datasets. The average length of the resulting sequence was approximately 22 m.

Top-1 accuracy was used as a performance index.

The KITTI dataset \cite{KITTI}  was obtained from a sensor-equipped vehicle in an urban environment in Karlsruhe, Germany. Formally, Visual Odometry/SLAM Evaluation 2012 in the KITTI dataset was used. An example of an image used is shown in Fig. \ref{fig:seven}. This dataset consists of 22 stereo sequences, 11 of which have acquired GPS coordinates, and the remaining 11 have no GPS coordinates. The camera uses PointGray Flea2 video cameras. The frame rate is 10 ps. The image resolution was modified from the original setting of 1,392$\times$512 pixels to that of the KITTI dataset of 1,241$\times$376 pixels.

\subsection{Results}

Table \ref{table:result} presents the performance results. The accuracy of the proposed method, which is based on the self-attention mechanism, was higher than that of any comparison method. Notably, ablation consisting of only the TPV module with the self-attention mechanism outperformed the baseline method of VGG16. In the early stages of the experiment, training was performed without using the weights pre-trained on ImageNet, and the accuracy was not high. However, fine-tuning of the weights pretrained on ImageNet significantly improved the accuracy.

Figure \ref{fig:successfailure} shows several success and failure examples for the self-attention-based TPV method. In the first failure example, the image was predicted as ``turn right". This may be because the gray-colored pixels on the right side part of the image were recognized as a road region. Notably, such a failure could be resolved by introducing the FPV module, which significantly improved the discriminativity between right and left turns. For the second example, the ground-truth intersection class was the right facing T-junction (class \#4), but in the failure example, it is predicted as a bottom facing T-junction (class \#6). This might be because the angle of the road on the left side of the T-junction is close to going straight. For the third example, the ground-truth class was crossroads (class \#7), but in the failure example, it was predicted as a right facing T-junction (class \#4). In this case, the road on the left was hidden behind a black car, and because of this, it was incorrectly classified right facing T-junction (class \#4). What is common to many examples is that the recognition often failed when the viewpoint location is distant from the center of the intersection. Figure \ref{fig:confusion} shows the statistics of the classification results in the form confusion matrix.

Currently, the I-Net is not trained end-to-end; instead, a handcrafted network is used. Notably, the performance of this handcrafted code is satisfactory in several situations. It often outperformed the machine learning variants we tried during the development process. However, failure modes were also observed. One such failure mode is the case in which the FPV module fails to predict the ego-motion because the rule prioritizes FPV decisions over TPV decisions. In other words, our framework may fail with high confidence. To avoid such a failure, statistics-based methods, such as neural networks or perceptron methods, would be effective and should be explored in future work.

\section{Conclusions}

In this study, we addressed the intersection classification problem from the perspective of the self-attention mechanism. This self-attention method was effectively incorporated into the unified FPV-TPV framework. Experiments verified that the self-attention-based TPV module is complementary to the FPV module and can improve the overall performance of the intersection classification.

\bibliography{reference} 
\bibliographystyle{unsrt} 

\editage{}{
\figL
\figM
\figD
\figF
\figK
\figP
\figT
}

\end{document}